\documentclass[conference]{IEEEtran}
\IEEEoverridecommandlockouts
\usepackage{amsmath,amssymb,amsfonts}
\usepackage{algorithmic}
\usepackage{graphicx}
\usepackage{textcomp}
\usepackage{xcolor}
\def\BibTeX{{\rm B\kern-.05em{\sc i\kern-.025em b}\kern-.08em
    T\kern-.1667em\lower.7ex\hbox{E}\kern-.125emX}}

\usepackage[numbers]{natbib}

\usepackage{hyperref}
\usepackage{float}
\usepackage{subcaption}
\begin{document}

\title{Anomaly Detection of Wind Turbine Time Series using Variational Recurrent Autoencoders}


\makeatletter
\newcommand{\linebreakand}{%
  \end{@IEEEauthorhalign}
  \hfill\mbox{}\par
  \mbox{}\hfill\begin{@IEEEauthorhalign}
}
\makeatother
\author{
\IEEEauthorblockN{Alan Preciado-Grijalva}
\IEEEauthorblockA{\textit{Hochschule Bonn-Rhein-Sieg}\\
\textit{Fraunhofer Center for Machine Learning and SCAI} \\
alan.preciado@h-brs.de}
\and
\IEEEauthorblockN{Victor Rodrigo Iza-Teran}
\IEEEauthorblockA{\textit{Fraunhofer Center for Machine Learning and SCAI} \\
victor.rodrigo.iza.teran@scai.fraunhofer.de}
}
\maketitle

\begin{abstract} 
Ice accumulation in the blades of wind turbines can cause them to describe anomalous rotations or no rotations at all, thus affecting the generation of electricity and power output. In this work, we investigate the problem of ice accumulation in wind turbines by framing it as anomaly detection of multi-variate time series. Our approach focuses on two main parts: first, learning low-dimensional representations of time series using a Variational Recurrent Autoencoder (VRAE), and second, using unsupervised clustering algorithms to classify the learned representations as normal (no ice accumulated) or abnormal (ice accumulated). We have evaluated our approach on a custom wind turbine time series dataset, for the two-classes problem (one normal versus one abnormal class), we obtained a classification accuracy of up to 96\% on test data. For the multiple-class problem (one normal versus multiple abnormal classes), we present a qualitative analysis of the low-dimensional learned latent space, providing insights into the capacities of our approach to tackle such problem. The code to reproduce this work can be found here \href{https://github.com/agrija9/Wind_Turbines_VRAE_Paper
}{https://github.com/agrija9/Wind-Turbines-VRAE-Paper}. 


\end{abstract}


\section{Introduction}

        

In engineering, it is of particular interest to study the freezing of the blades of wind turbines caused by harsh weather conditions. In order to ensure the optimal operation of such systems, we can analyze and gain meaningful insights from the data they generate. For this, certain data manipulation tasks like compression, clustering, and reconstruction are helpful.

Data clustering, for example, attempts to visually identify data patterns in low dimensional spaces. Numerous physical spatio-temporal insights can be derived from this, since clustering in low-dimensional spaces indicates data has (un)correlated attributes. 

With more data available and novel machine learning models being developed, it is reasonable to explore their potential to tackle the problem of ice accumulation in wind turbines. Suitable models for these tasks are Variational Autoencoders (VAEs)\cite{Kingma_2013} due to their capacities to compress, generate and project data into a low-dimensional space. VAEs have achieved state-of-the-art results on image generation \cite{Jaskari_2018}, clustering and anomaly detection \cite{Pereira_I_2019}\cite{Pereira_II_2019}\cite{Zhao_2018}, and data reconstruction \cite{Wetzel_2017}. 

In this work, we report on the implementation of a fully unsupervised learning pipeline for anomaly detection in the blades of wind turbine time series simulations data. Our pipeline consists of two main steps: (1) learning abstract low-dimensional time series data representations with a Variational Recurrent Autoencoder (VRAE) \cite{Fabius_2014} and (2) classifying these representations using clustering algorithms (KMeans++, Hierarchical Clustering and DBSCAN). 
Great emphasis has been put on the first part, since fine-tuning a VRAE neural network model requires extensive experimentation. 


\section{Related Work}



Dimension reduction methods like PCA and kernel-PCA \cite{Scholkopf_1997} have been effective in many use cases and straightforward to implement \cite{Nguyen_2019}. However, a numerous amount of datasets contain non-linearities that can not be captured by these methods, thus having to resort to other non-linear techniques like t-SNE \cite{van_Der_Maaten_2008}. Just like PCA, t-SNE is also used for visualizing large datasets, it uses a random walk on neighborhood graphs to reveal structure at different scales. These methods of dimension reduction are a key tool to gain insights into any clustering behavior datasets may present. Furthermore, highly non-linear neural network approaches have shown promising results, in the case of time series dimensionality reduction and clustering, VAEs have achieved state-of-the art results on anomaly detection \cite{Pereira_I_2019}\cite{Zhao_2018}. \cite{Pereira_I_2019} demonstrated state-of-the-art accuracy for the detection of anomalies in the \href{http://www.timeseriesclassification.com/description.php?Dataset=ECG5000}{ECG500 dataset} with an accuracy exceeding 90\%.

Moreover, \cite{Zhao_2018} extracted the relationship between time series variables
obtained from the monitoring of wind turbine systems. This group worked with an autoencoder network based on Restricted Boltzmann Machines, successfully implementing an early warning of faulty components and deducing the physical location of such components.

\section{Proposed model for anomaly detection}

        
    The proposed model for ice detection in the rotor blades consists of two fundamental steps: unsupervised representation learning and anomaly detection. 

\subsection{Unsupervised representation learning}
    The model that we use for this task is the Variational Recurrent Autoencoder (VRAE). More formally, let $\chi=[\textbf{x}^{(n)}]^N_{n=1}$ be the time series dataset composed of $N$ sequences, with each sequence having a length $T$, $x^{(n)} = [x^{(n)}_1, x^{(n)}_2,...,x^{(n)}_T]$, and each data point $x_t^{(n)}$ is a $d_x$ dimensional vector (number of features).
    
    The encoder of the VRAE takes each time series $x^{(n)}$ and it is parametrized by a long short-term memory (LSTM) layer that at each time step computes a hidden state $h_t^{enc}$. The last hidden state $h_T^{enc}$ is thus an abstract representation that represents the whole given sequence \textbf{x}. Similarly to \cite{Pereira_I_2019}\cite{Pereira_II_2019}, the prior distribution $p(\textbf{z})$ is a multi-variate normal distribution $\mathcal{N}$(\textbf{0}, \textbf{I}). The parameters that approximate the posterior distribution $q_\psi(z|x)$, $\mu_z$ and $\Sigma_z$, are obtained by taking mean and standard deviation from this last hidden state by using two fully connected layers with a SoftPlus activation. According to [25], using a SoftPlus activation ensures that variance is non-negative . 
    The latent variables \textbf{z} are sampled from the parametrized posterior $q_\psi(z|x)$ via $\mu_z$ and $\Sigma_z$ by using the re-parametrization trick by doing 
    
    \begin{equation}
    z = \mu_z + \sigma_z \odot \epsilon
    \end{equation}
    
    Where $\epsilon \sim \mathcal{N}(\textbf{0}, \textbf{I})$ is Gaussian noise and $\odot$ corresponds to element-wise product.
    
    The decoder of the VRAE is another LSTM network that takes as input the latent vector \textbf{z} from the approximate posterior and outputs at each time step $t$ the parameters that reconstruct the input variable \textbf{x}. Similar to the encoding distribution, the decoding distribution $p_{\phi}(x|z)$ is defined as a multi-variate Gaussian distribution.
     The loss function is the VAE loss function introduced in \cite{Kingma_2013} and the training procedure follows subsequently following gradient update and stochastic gradient descent.\\
     
\subsection{Anomaly detection}
    We perform anomaly scoring using the learned low-dimensional time series representations provided by the VRAE model. Following the procedure of \cite{Pereira_I_2019}, the model is mapping sequences \textbf{x} into a lower-dimensional space and we then project them into two dimensions using PCA and t-SNE in order to evaluate grouping in specific regions. This makes it more feasible for a clustering method to detect normal vs anomalous (abnormal) cases. 
    Anomaly detection consists, therefore, in detecting if a latent representations is normal or abnormal. In this work, we have implemented this detection using clustering algorithms. 
    
    Clustering algorithms give a numerical label to each latent representation, framing the problem as a two-class or multiple-class classification problem. We are taking this approach based on the fact that the model is capable of learning representations that tend to group in lower-dimensions given a balanced normal and abnormal percentage of training data, also assuming there is statistical difference between cases. We have applied three different clustering methods in the representations: \textit{k-means ++} \cite{Arthur_2007}, density-based spatial clustering (\textit{DBSCAN}) \cite{Ester_1996} and \textit{hierarchical clustering} \cite{reynolds_2006}. These methods are set to find two (or more) clusters given the number of classes (normal and abnormal). The output is then matched to the ground truth labels given corresponding to the actual class each representation belongs to. With this, we can compute a classification accuracy.

\section{Experiments and Results}

    In this section we describe the wind turbine dataset used in our experiments, relevant data preprocessing steps, and the results obtained for the case of two and multiple classes representation learning and anomaly scoring.
    
    \subsection{Dataset Description}

        Our dataset is composed of wind turbine multivariate time series simulated with an open source whole-turbine simulator called \href{https://github.com/OpenFAST/openfast}{FAST}. We model and collect time series data of a wind turbine operating with/without ice on the rotor blades. We categorize ice accumulation in one blade into three zones: the first zone covers the first half and the two other zones divide the second half of the blade into two again. For each simulation, each one of these three zones contains a particular ice mass. The convention to refer to the region and amount of ice mass is $x-y-z$. For example, a configuration $0.4-0.6-0.8$ implies 0.4 kg of ice in zone one, 0.6 kg in zone two and 0.8 kg in zone three.
    
        As an initial approach, we have filtered the dataset to contain only simulations with ice mass in one zone at a time (i.e. no ice in two other zones). This is to investigate as a first approximation the capacity of our model to cluster time series in lower dimensions based on ice mass configuration in one zone at a time.
        
        To setup an anomaly detection problem, we define specific classes of times series. A \textit{normal} time series corresponds to the configuration $0-0-0$ (no ice in any zone), an \textit{abnormal} time series corresponds to any configuration in eq. \eqref{eq:2}. 
        
        \begin{equation} 
    		\begin{split}\label{eq:2}
    		zone1  \rightarrow x_{mass} - 0 - 0 \\
    		zone2  \rightarrow 0 - y_{mass} - 0 \\
    		zone3  \rightarrow 0 - 0 - z_{mass}
    		\end{split}
    	\end{equation}
    	
    \subsection{Data Preprocessing}
    	We have balanced the dataset to have approximately the same number of normal and abnormal time series. It is composed of 14 normal and 11 abnormal simulations. Each simulation corresponds to a time series consisting of 10,000 time steps and 27 sensor features in total. 
        
        
        First, we MinMax scale the data to a range [-1 to 1], this is a crucial preprocessing step since the learning performance of the model can be affected by the various sensor amplitudes recorded. Second, we select only 6 out of the the initial 27 features. These filtered features are the accelerations in flap-wise and edge-wise components for three blades (see table \ref{table:features}). We have chosen these features based on the idea that this physical information is sufficient for the model to be able to cluster time series in lower dimensions efficiently. 
        Fig.\ref{fig:wind_turbine_dataset} illustrates a sample simulation after the scaling and feature filtering steps.
        
        \begin{table}[ht]
     	 	\caption{Filtered features from the wind turbine simulations.}
        	\label{tab:table1}
    	    \begin{tabular}{ |p{1.5cm}||p{6.4cm}|  }
    		 \hline
    		 \hline
    		 Parameter & Description\\
    		 \hline
    		 Spn1ALxb1 & Blade 1 flapwise acceleration (absolute) span station 1 \\
    		 Spn1ALyb1 & Blade 1 edgewise acceleration (absolute) span station 1  \\
    		 Spn1ALxb2 & Blade 2 flapwise acceleration (absolute) span station 1 \\
    		 Spn1ALyb2 & Blade 2 edgewise acceleration (absolute) span station 1 \\
    		 Spn1ALxb3 & Blade 3 flapwise acceleration (absolute) span station 1\\
    		 Spn1ALyb3 & Blade 3 edgewise acceleration (absolute) span station 1\\
    		 \hline
    		\end{tabular}
    	\label{table:features}
    	\end{table}
    	
    	Next, we reshape the data to the format \textit{(samples, timesteps, features)} since this is the required input format for LSTM blocks that compose the VRAE. Finally, we slice each time series into smaller segments of 200 to 1000 time steps. This is because our simulations are set up such that we have roughly 12 rotations per minute, this means there is one rotation every 5 seconds, this corresponds to a length of approximately 1000 time steps, on top of this, smaller time steps avoid vanishing gradients. In summary, our procedure generates a total of 1250 time series. 
    	
    	\begin{figure}[H]
    	    \hspace*{-0.3cm}
    	    \includegraphics[scale=0.23]{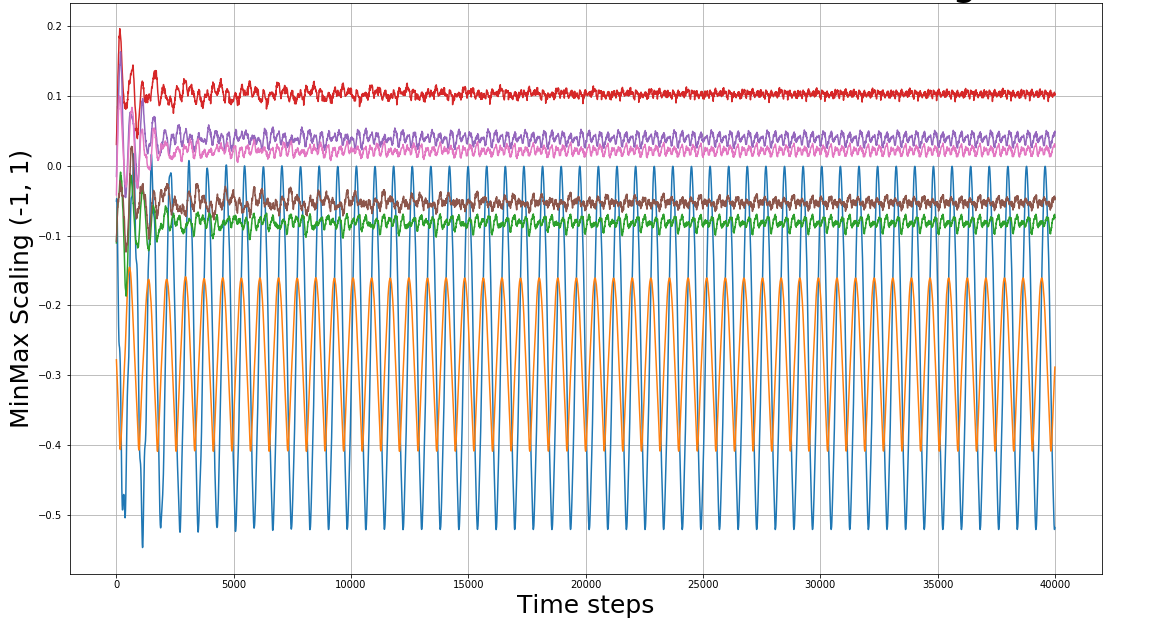}
    	    \caption{A simulated time series from our dataset. Each color line corresponds to a specific physical variable recorded.}
    	    \label{fig:wind_turbine_dataset}
    	\end{figure}
    
    \subsection{Anomaly detection with two classes}
    
    The first problem we tackled is the detection of anomalies given two classes. We have taken normal time series ($0-0-0$) and one abnormal time series (zone 1:  $x_{mass} - 0 - 0$).
    
    Following the proposed model for ice detection mentioned in Section III, we trained a VRAE using a single hidden LSTM layer with 90 units (one hidden layer gave better results than multiple hidden layers). We used the Adam optimizer \cite{kingma_2014}, gradient clipping (to avoid gradient explosion) and dropout in the hidden LSTM layer with rate of $0.2$. A bottleneck layer (variational layer) maps these $90$-dimensional vectors into $20$-dimensional vectors (i.e. latent space dimensions). We used a learning rate of $l=0.0005$ and momentum $0.9$. The data was split in 70\% training and 30\% validations sets with time series split into chunks of 200 time steps and the 6 features described in table \ref{table:features}. We loaded data into memory in batches of 64 and performed training for 2000 epochs. The reconstruction term in the loss objective is the mean squared error (MSE loss) given that we assume that the parametrized posterior probability $p(x|z)$ is a normal distribution. The total training time was approximately 30 minutes using an NVIDIA Tesla v100 GPU with 32GB of RAM.
    
    \subsubsection{Analysis of latent space}
    
        Before training the VRAE, we ran a sanity check by applying PCA directly to the test set to verify if this method can linearly separate the data. We observed that pure PCA is not good enough to cluster normal and abnormal time series in distinctive regions, indicating that such techniques can not capture non-linearities that exist in our data. 
        
        Fig.\ref{fig:2d_latent_space_two_classes} shows the results obtained on the test set after training the VRAE. The data points correspond to a projection of the 20-dimensional latent vectors into 2-dimensional vectors using PCA (first and second components), t-SNE and Spectral Embedding. We note that in most cases, the data points are properly clustered according to normal (red points) and abnormal (blue points) classes in specific regions of the 2D plane. 
        The location of the 2D data points depends on each projection method, but the underlying clustering behavior is quite evident in each case, thus proving that the model has learned to identify when a time series contains ice or no ice. From here, we highlight the importance of learning good abstract representations of time series, since having such distinctive grouping behavior can make unsupervised clustering algorithms perform better. We have colored the 2D data points to have a visual reference since we know the ground truth labels for each of them, however, the training and projections are fully unsupervised. 
        
        \begin{figure*}[ht!]
          \centering
          \begin{subfigure}{0.321\linewidth}
            \includegraphics[width=\linewidth]{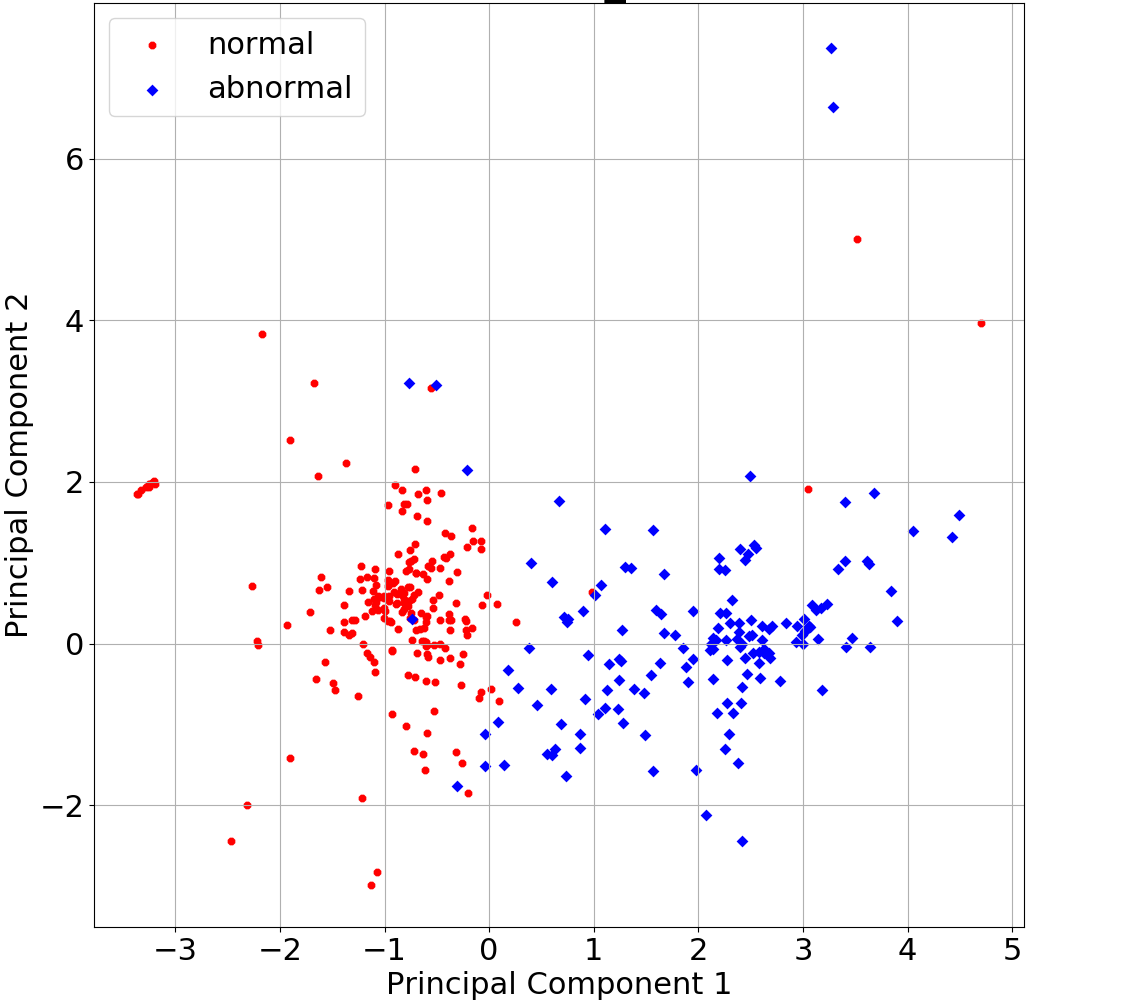}
            \caption{PCA}
          \end{subfigure}
          \begin{subfigure}{0.325\linewidth}
            \includegraphics[width=\linewidth]{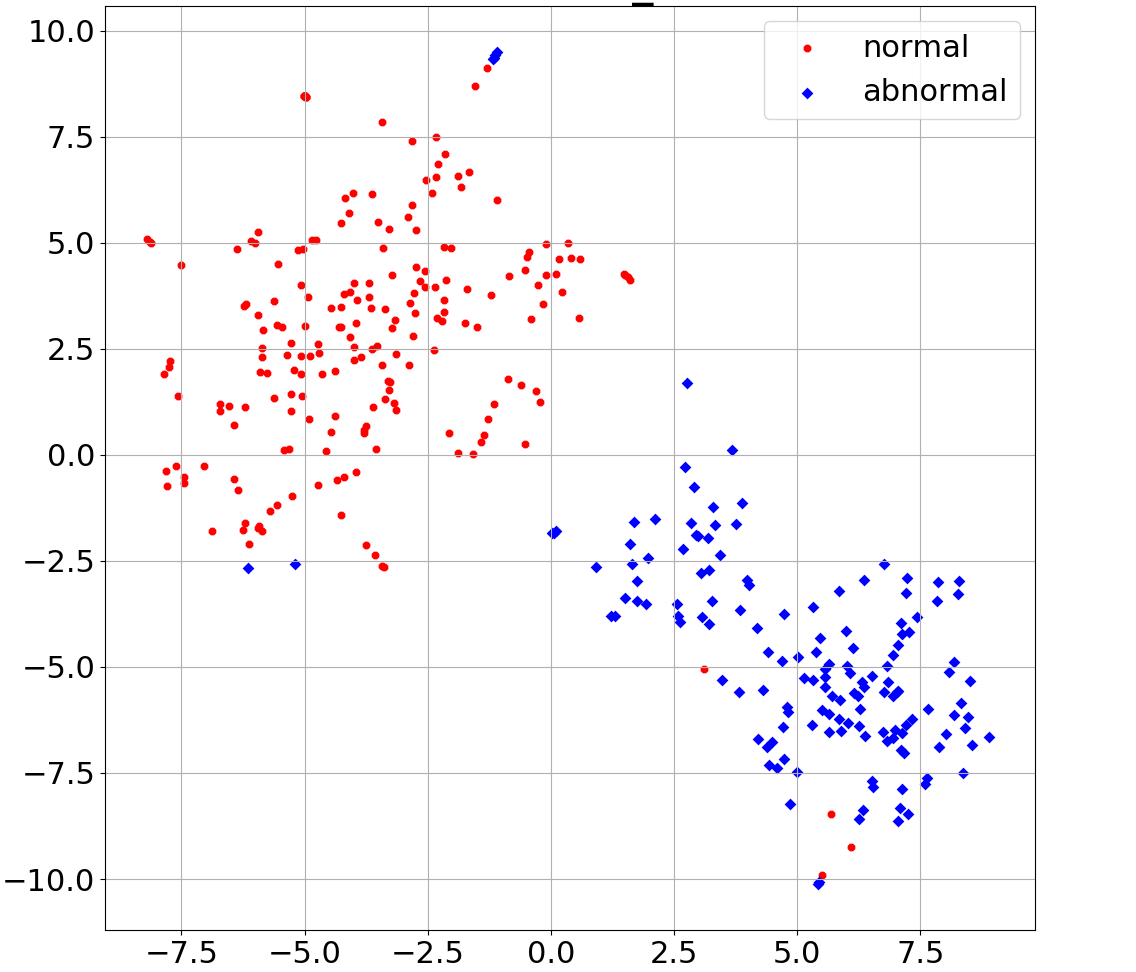}
            \caption{t-SNE}
          \end{subfigure}
          \begin{subfigure}{0.325\linewidth}
            \includegraphics[width=\linewidth]{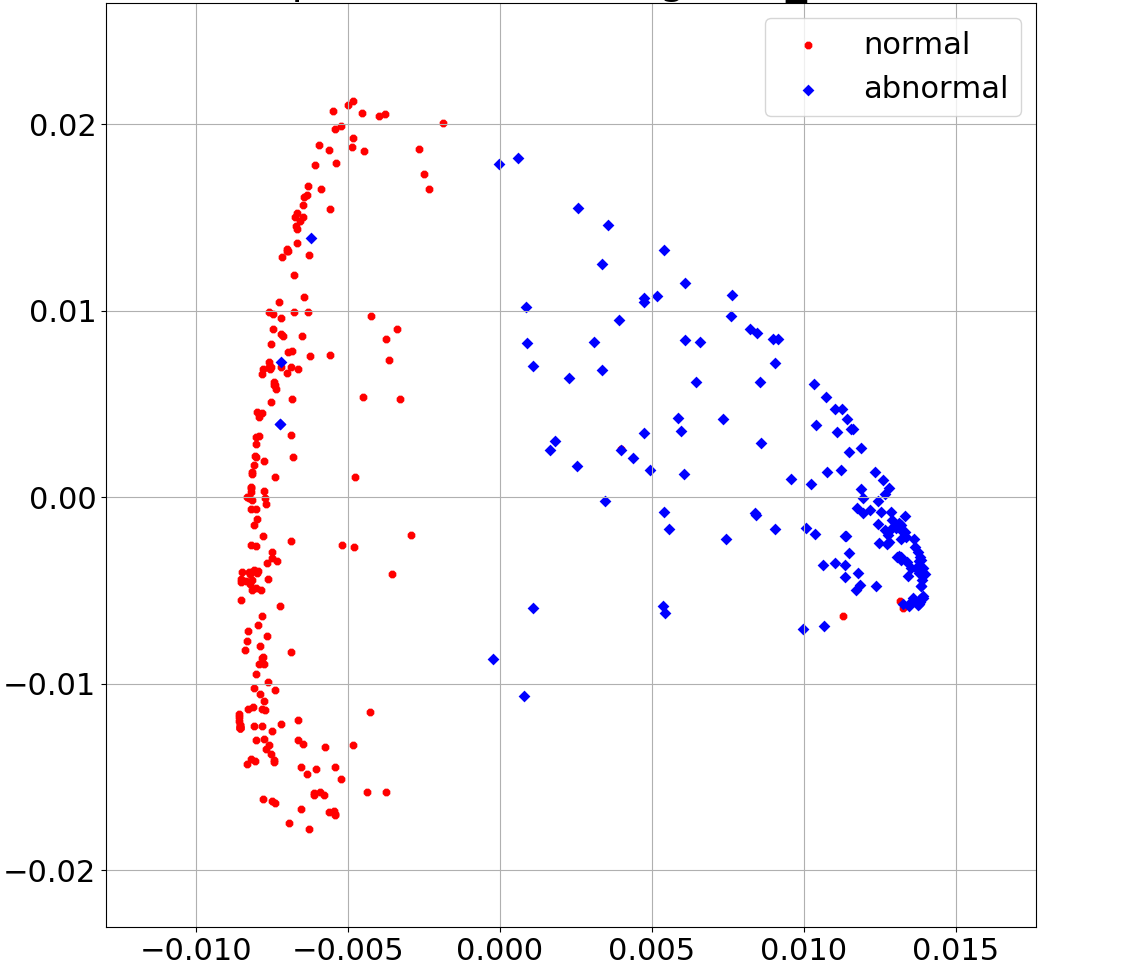}
            \caption{Spectral Embedding}
          \end{subfigure}
          \caption{2D projections of the 20-dimensional representations obtained with the VRAE.}
          \label{fig:2d_latent_space_two_classes}
        \end{figure*}
        
        
        In order to further analyze the learned representations by our model, we took 15 samples of normal and abnormal 20-dimensional latent vectors and plot each dimension horizontally (without projecting them into 2D). 
        
        In Fig.\ref{fig:latent_vectors_as_lines}, the red lines correspond to normal latent vectors and the blue lines to abnormal ones. Here, the x-axis corresponds to each entry of the 20-dimensional vectors and the y-axis are the normalized amplitudes provided directly by the model. Remarkably, we can see that even in their 20-dimensional representation, red lines tend to describe different maxima and minima compared to blue lines. Note for instance, that the dimensions 1, 5, 7, 13 and 19 are the ones where normal and abnormal classes have the less correlation, knowing what dimensions provide the highest contrast between our classes can in principle allow us to further reduce the dimensions of the latent vectors from 20 to 5 or 6.
    
        The next step in our pipeline consists in applying clustering algorithms on top of the projected 2D data points. These models classify data based on a predefined number of clusters thus providing a final anomaly scoring.
        
        \begin{figure}[H]
    	    \centering
    	    \includegraphics[scale=0.30]{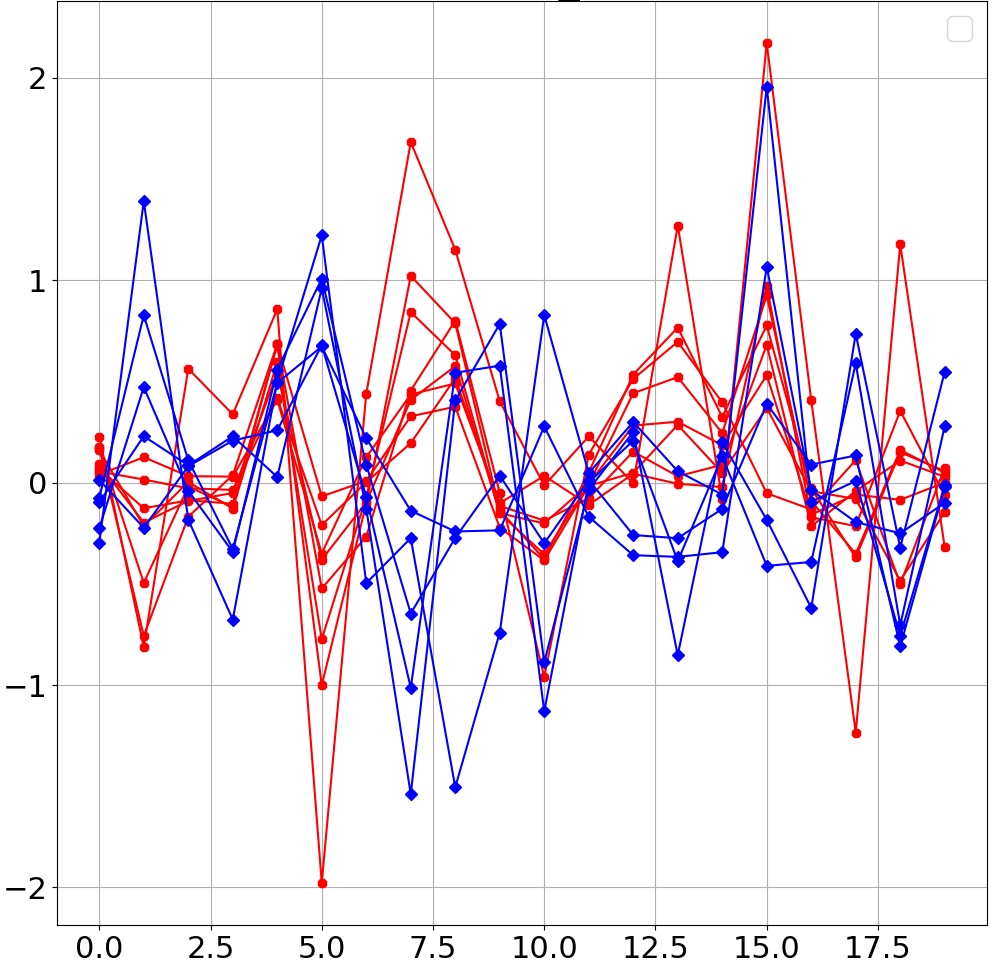}
    	    \caption{VRAE 20-dimensional latent vectors plotted as horizontal lines. Red lines are normal classes, blue lines are abnormal classes.}
    	    \label{fig:latent_vectors_as_lines}
    	\end{figure}
    
    \subsubsection{Unsupervised clustering and anomaly scoring}
    	
    	Fig.\ref{fig:unsupervised_clustering_two_classes} shows the clustering results of KMeans++ and Hierarchical clustering on our 2D data points (obtained with PCA). Both algorithms have been set to identify data points between two clusters. In general, we see that both methods perform efficient clustering when compared to the groundtruth data labels (figure on the left). Furthermore, the advantage of these methods is that they are relatively fast and straightforward to implement.
    	
        
        \begin{figure*}[ht!]
          \centering
          \begin{subfigure}{0.32\linewidth}
            \includegraphics[width=\linewidth]{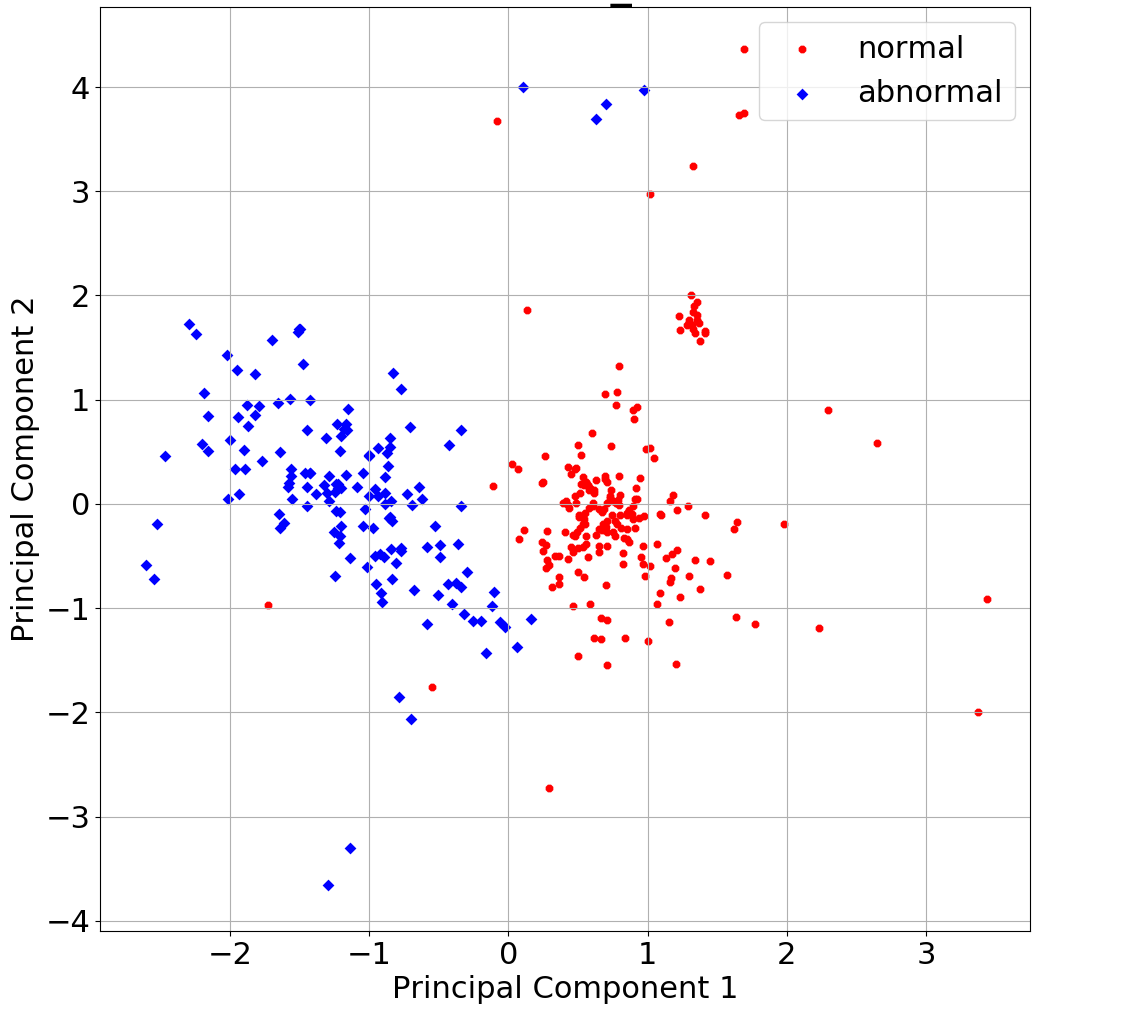}
            \caption{Groundtruth}
          \end{subfigure}
          \begin{subfigure}{0.33\linewidth}
            \includegraphics[width=\linewidth]{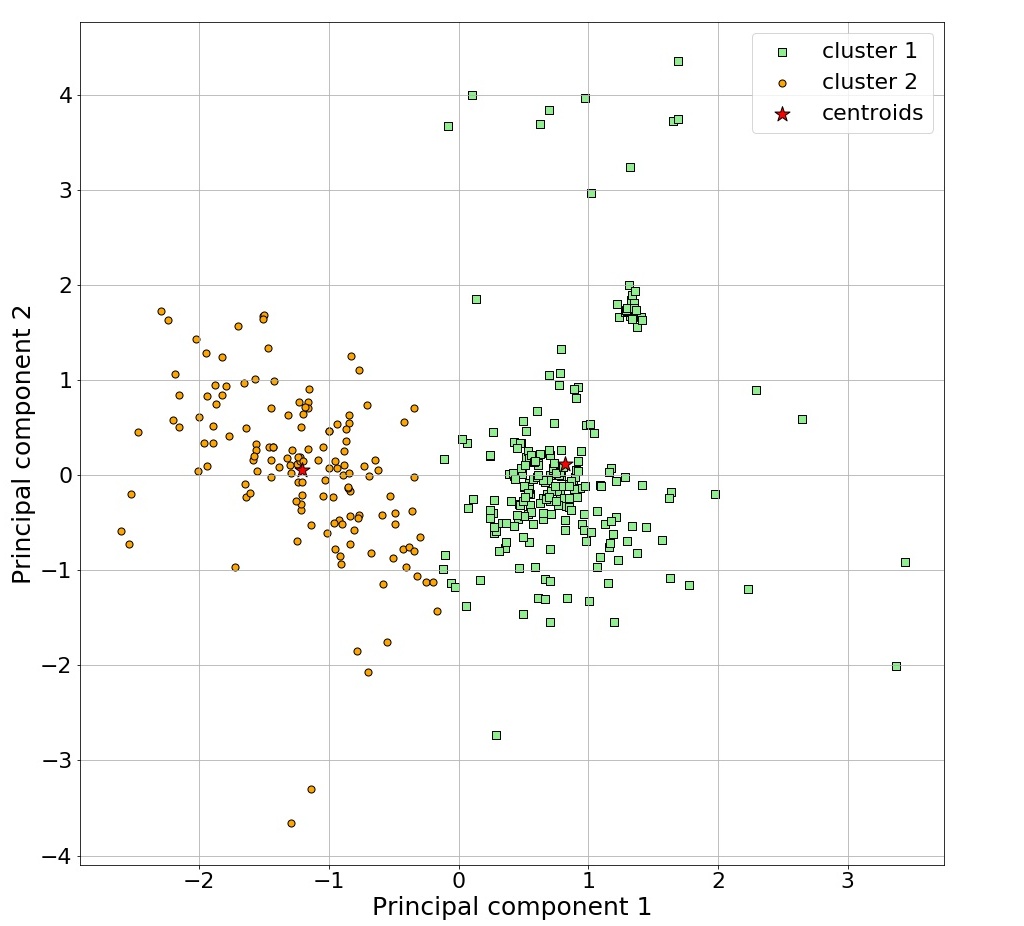}
            \caption{KMeans++}
          \end{subfigure}
          \begin{subfigure}{0.33\linewidth}
            \includegraphics[width=\linewidth]{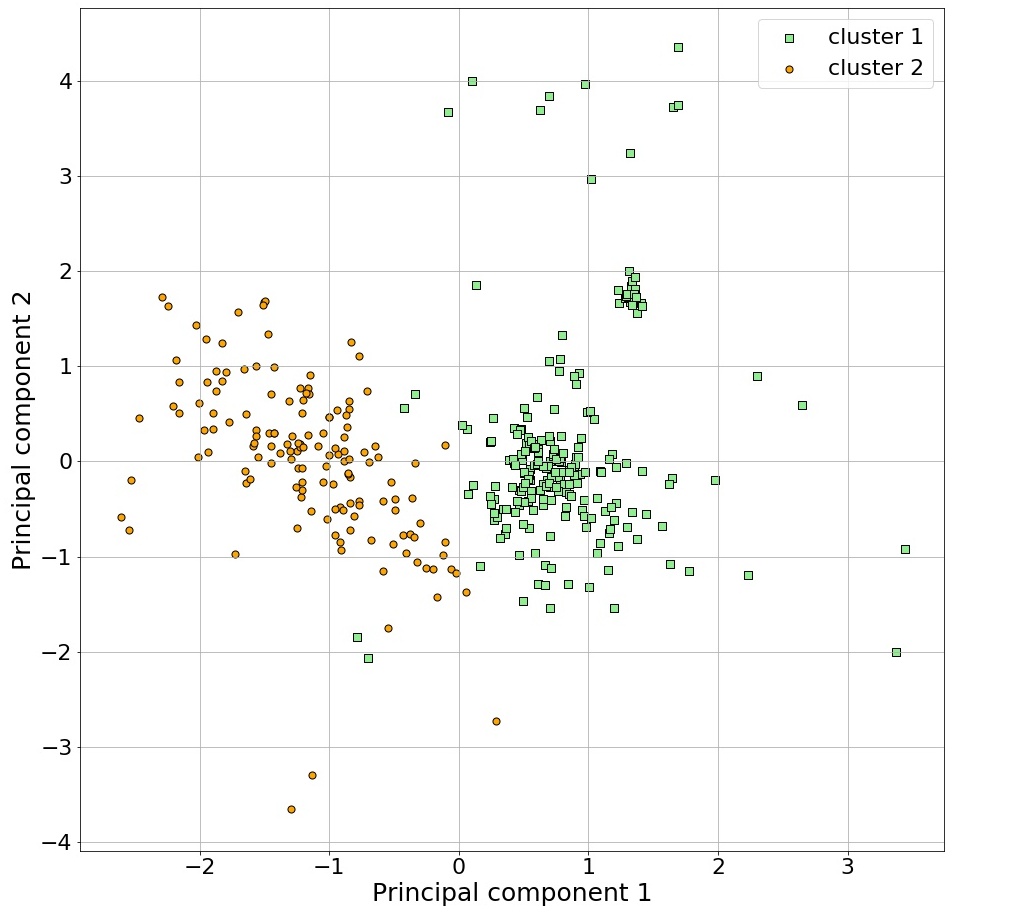}
            \caption{Hierarchical Clustering}
          \end{subfigure}
          \caption{Visual evaluation of clustering performance of KMeans++ and Hierarchical Clustering. Both methods label most data points correctly when compared to the groundtruth labels.}
          \label{fig:unsupervised_clustering_two_classes}
        \end{figure*}
        
        In order to provide a quantitative evaluation for anomaly scoring, we have computed several metrics like classification accuracy, area under the curve (AUC) and precision. We have added a third clustering method (DBSCAN) for a broader evaluation. Table \ref{table:anomaly_scoring} summarizes the obtained results.
    	
    	\begin{table}[h]
    		\centering
     	 	\caption{Anomaly scoring results using 3 unsupervised clustering algorithms.}
        	\label{tab:table1}
    	    \begin{tabular}{ |p{1cm}||p{2cm}||p{2cm}||p{2cm}|}
    		 \hline
    		 \multicolumn{4}{|c|}{\textbf{Anomaly scoring results}} \\
    		 \hline
    		 Metric & KMeans++ & Hierarchical & DBSCAN\\
    		 \hline
    		 Accuracy & \textbf{0.9667} & 0.9639 & 0.6315\\
    		 AUC & \textbf{0.9619} & 0.9605 & 0.5507 \\
    		 Precision & 0.9674 & 0.9641 & \textbf{0.9890}\\
    		 Recall & \textbf{0.9667} & 0.9639 & 0.6315 \\
    		 F1-score & \textbf{0.9666} & 0.9638 & 0.7616\\
    		 \hline
    		\end{tabular}
    		\label{table:anomaly_scoring}
    	\end{table}
    	
    	The results from table \ref{table:anomaly_scoring} indicate that DBSCAN has a lower performance overall. 
    	The other two methods on the contrary, achieved classification accuracies of up to 96\%. 
    	These classification results indicate that our anomaly detection proposed framework is being able to classify between normal and abnormal time series up to 96\% of the time. 
    
    \subsection{Unsupervised time series representation learning for multiple classes}

    	The next problem we tackled is anomaly detection for multiple classes. Here, we have taken into account all three zones to compose the training dataset (i.e. weights in zone 1, zone 2 and zone 3). We trained the VRAE with a single hidden LSTM layer, in this case, we modified this layer to 128 units. The bottleneck layer projects these 128 dimensional representations into latent vectors of 5 dimensions (we changed from 20 to 5 after tuning this parameter). Adam optimizer is used, gradient clipping (to avoid gradient explosion) and dropout in the hidden layer with rate of 0.2. The learning rate is $l=0.0005$ and momentum $0.9$. The data is split in 70\% training and 30\% validations sets with time series split into chunks of 200 time steps and the 6 features described in the previous chapter. We load batches of 64 into memory and perform training for 2000 epochs. The reconstruction term in the loss objective is the MSE loss. The training time is approximately 1.5 hours. \\
        
        \subsubsection{2D latent space of time series for multiple classes}
            
            Fig.\ref{fig:2D_projections_multiple_classes_balanced_dataset} shows the 2D projections of the 5-dimensional latent vectors of t-SNE and spectral embedding after training the model with a balanced number of classes (around 600 time series per class). In this case, we obtained that the model is capable of clustering normal versus abnormal cases successfully (red dots versus the rest), however, it is not capable of clustering abnormal classes depending on their zone (i.e. blue, green, black dots are not clustered). To investigate this behavior, we have plotted a few 5-dimensional vectors from all three abnormal samples (similar to the analysis in Fig.\ref{fig:latent_vectors_as_lines}), we have seen that there is not a lot of variability between lines of abnormal zones themselves, they rather tend to describe the same maxima and minima. We assume that this occurs due to the fact that training with a fixed balanced number of samples per class prevents the model to regularize the latent space properly.
            
            


            We ran a series of experiments to overcome the problem of clustering abnormal zones correctly: First, we implemented a VRAE using bi-directional LSTMs and increased the number of timesteps of each time series from 200 to 500. However, for the bi-directional LSTM we obtained similar results as the ones shown in Fig.\ref{fig:2D_projections_multiple_classes_balanced_dataset} (with uni-directional LSTM), in the case of longer time steps, we obtained even worse results, this can have to do with the fact that longer sequences implies less available training samples. 
            
            As a second experiment, we increased the number of abnormal samples for training (in previous experiments, the training set was balanced, but since there are more abnormal samples available, we have trained the model with all the available data), in addition to this, we implemented cyclical annealing to mitigate vanishing of the KL divergence \cite{Hao_Fu_2019}. After training the model we projected the test set into 5-dimensional latent vectors and then into 2D using t-SNE and Kernel PCA (radial basis function), the results are shown in Fig.\ref{fig:2D_projections_multiple_classes_full_dataset}. From these projections, we can see a more successful formation of clusters. We note in both projections that time series from abnormal zone 1 (blue dots) form a very distinctive cluster, zones 2 and 3 (green and black dots) have some overlapping but they still present clustering. These results indicate that the model is capable of clustering time series from all abnormal zones and anomaly scoring can be applied next.

            \begin{figure}[H]
              \begin{subfigure}{\linewidth}
                \hspace*{0.38cm}
                \includegraphics[scale=0.3]{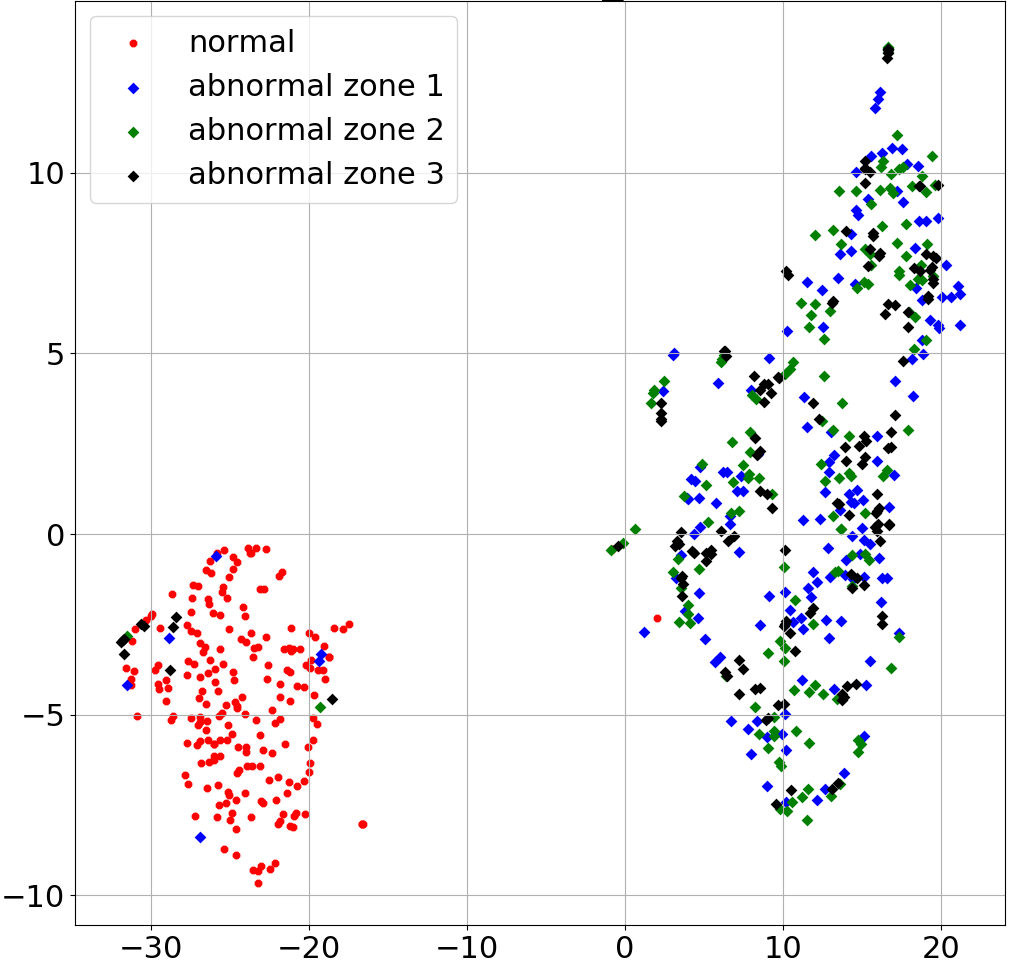}
                \caption{t-SNE projections}
              \end{subfigure}
              \begin{subfigure}{\linewidth}
                \includegraphics[scale=0.3]{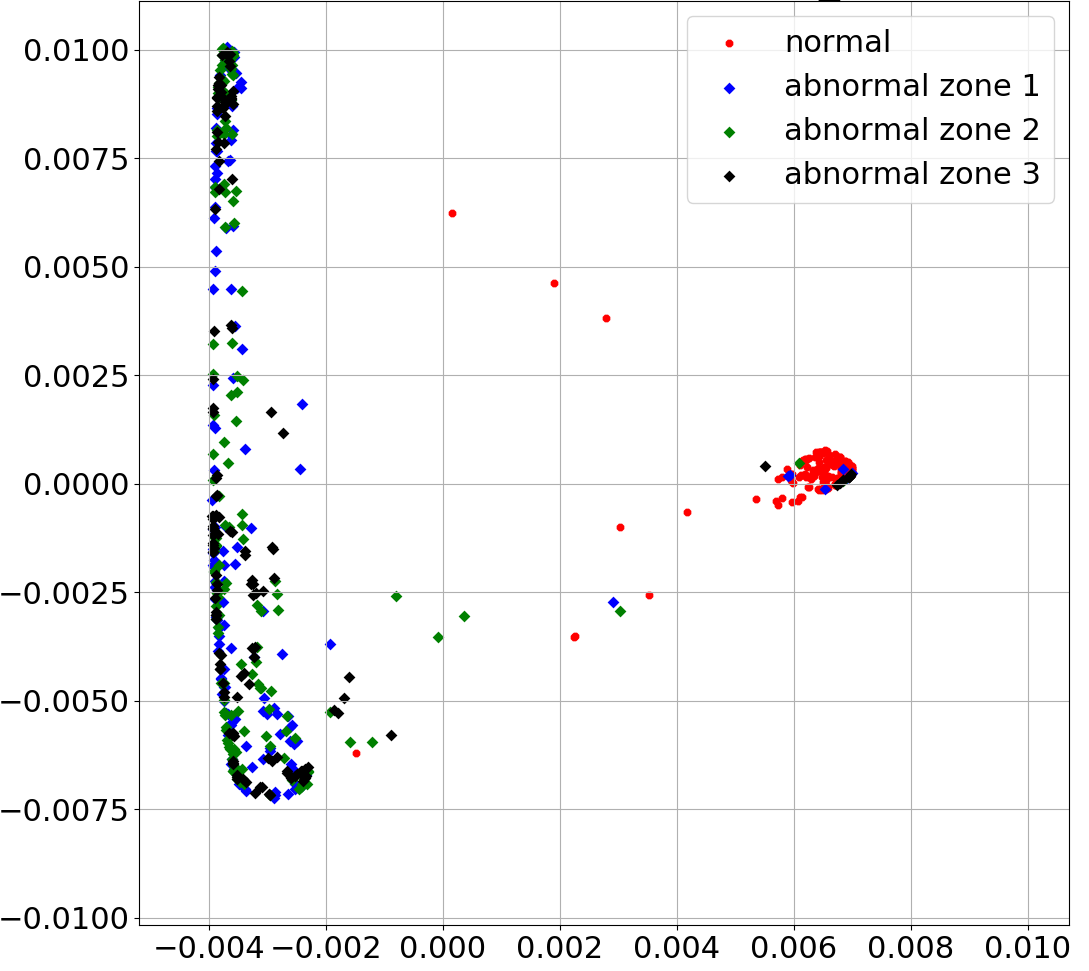}
                \caption{spectral embedding projections}
              \end{subfigure}
              \caption{Test set 2D projections of 5-dimensional latent vectors after training with all 3 classes (balanced dataset).}
              \label{fig:2D_projections_multiple_classes_balanced_dataset}
            \end{figure}
        
        	\begin{figure}[H]
              \begin{subfigure}{\linewidth}
                \includegraphics[scale=0.3]{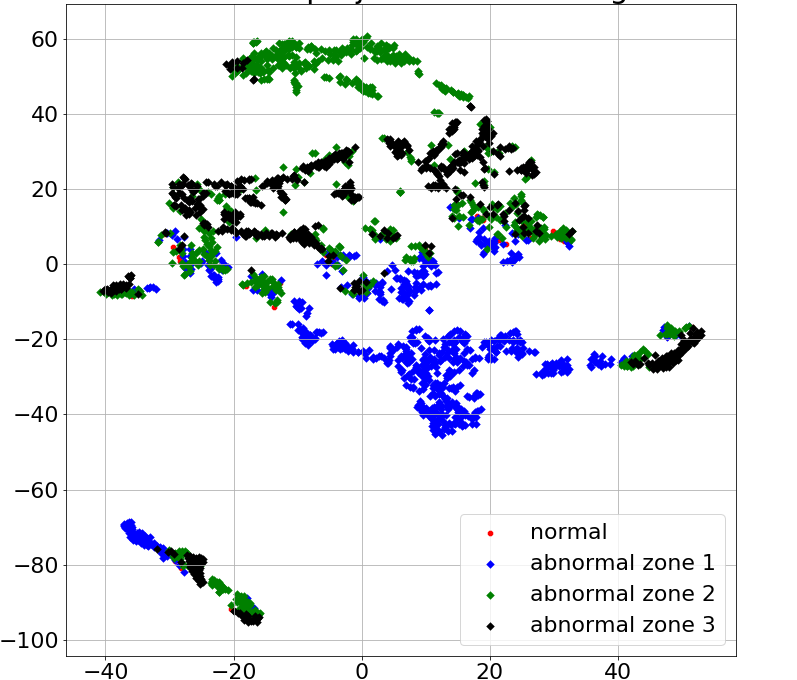}
                \caption{t-SNE projections}
              \end{subfigure}
              \begin{subfigure}{\linewidth}
                \includegraphics[scale=0.3]{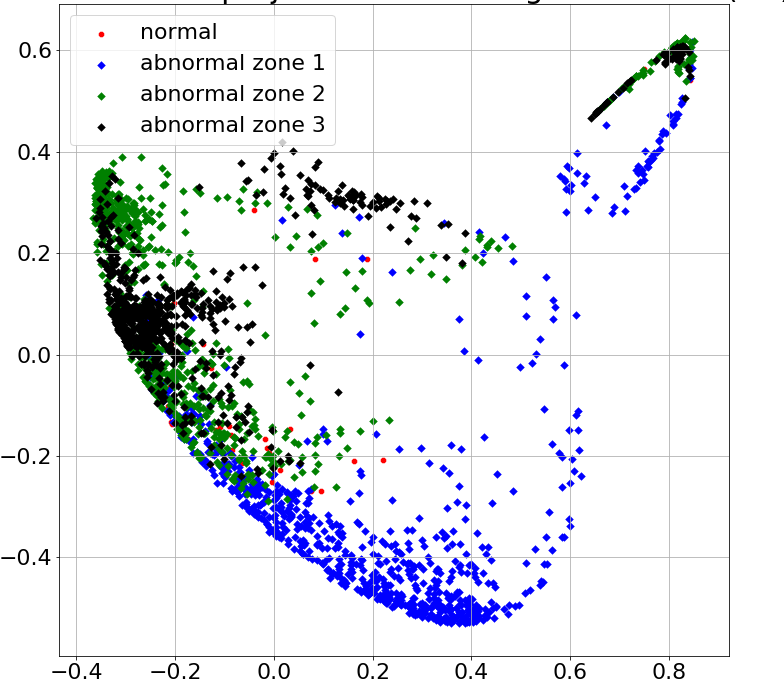}
                \caption{Kernel PCA}
              \end{subfigure}
              \caption{Test set 2D projections of 5-dimensional latent vectors after training with full dataset.}
              \label{fig:2D_projections_multiple_classes_full_dataset}
            \end{figure}

\section{Conclusions}
        In this work, we have introduced an end-to-end unsupervised learning pipeline for wind turbine anomaly detection; first, we have built a dataset composed of wind turbine multi-variate time series and preprocessed this data to train a Variational Recurrent Autoencoder, secondly, we have used this model to learn low-dimensional latent representations and implemented clustering algorithms on top of these latent representations to predict class correspondences. In the case of anomaly detection for two classes, we have obtained a classification accuracy of 96\% on the test set, furthermore, our approach allows us to draw intuitive conclusions and improves interpretability due to the fact that our analysis is two dimensional. In the case of anomaly detection for multiple time series classes, we have observed that the amount of training data plays an important role; we showed that our model is capable of clustering time series better when more data samples are available, these results take us one step further for efficient representation learning for anomaly scoring in the case of multiple classes. From our insights, we conclude that learning high-quality data representations with neural network is a key aspect of our approach.

\bibliographystyle{plainnat}
\bibliography{bibliography.bib}


\end{document}